\documentclass[11pt]{article}

\usepackage[preprint]{acl}

\usepackage{times}
\usepackage{latexsym}

\usepackage[T1]{fontenc}

\usepackage[utf8]{inputenc}

\usepackage{microtype}

\usepackage{inconsolata}

\usepackage{graphicx}

\usepackage{booktabs}

\usepackage{graphicx}
\usepackage{subcaption} 
\usepackage{xspace}
\usepackage{multirow}
\usepackage{amsmath}
\usepackage{amssymb}

\newcommand{\ours}{\textsc{OutlineForge}\xspace}

\title{\ours: Hierarchical Reinforcement Learning with Explicit States for Scientific Writing}

\author{
Yilin Bao \\
UC San Diego \\
yibao@ucsd.edu
\And
Ziyao He \\
Ohio State University \\
ziyaohe@ohio.edu
\And
Zayden Yang \\
Sheltered.AI \\
zayden.yang@sheltered-ai.edu
}

\begin{document}
\maketitle
\begin{abstract}
Scientific paper generation requires document-level planning and factual grounding, but current large language models, despite their strong local fluency, often fail in global structure, input coverage, and citation consistency. We present a reinforcement learning framework that casts scientific outline construction as a long-horizon planning problem over hierarchical document structures. Our approach models edit evolving outlines through structured actions, enabling the system to incrementally build a complete scientific manuscript. To support effective and stabilize learning,we introduce a two-stage optimization procedure consisting of (i) backward outline reconstruction from partial plans to enforce global structural consistency, and (ii) forward value-guided reinforcement learning with rewards explicitly modeling scientific correctness, discourse coherence, and citation fidelity. In addition, We further introduce a benchmark for scientific paper generation that evaluates document planning, input utilization, reference faithfulness, outline organization, and content-level factual accuracy. Our results show consistent improvements over strong neural and LLM baselines, particularly in long-range structural coherence and citation reliability.
\end{abstract}

\section{Introduction}

The scientific paper generation task is the intersection of long-form text generation,document-level planning, and verifiable scientific reasoning. Compared with short form tasks, it requires models not only to write fluently, but also to maintain coherent global structure, cover relevant inputs, and use citations in a faithful and consistent manner.\\

Over the past two years, large language models (LLMs) have undergone a rapid shift from passive text generators toward increasingly agentic systems capable of structured reasoning, tool use, long-horizon planning, and autonomous problem solving. Advances in model architectures, post-training methods, and interaction frameworks have enabled LLMs to perform tasks that require not only linguistic fluency but also procedural competence, verifiable reasoning, and iterative decision making. This evolution has fueled a growing body of research that views LLMs less as static predictors and more as general-purpose cognitive engines with the potential to participate in complex scientific and engineering workflows.\\

Parallel to these developments, artificial intelligence has become deeply embedded across the scientific research pipeline. Surveys such as AI4Research~\cite{chen2025ai4researchsurveyartificialintelligence} and The Evolving Role of Large Language Models in Scientific Innovation~\cite{zhang2025evolving} document how AI systems now operate as literature reviewers, experiment planners, hypothesis evaluators, and even autonomous research collaborators. Within this emerging paradigm, LLMs are no longer just tools that accelerate human workflows; they increasingly take on structured roles in knowledge discovery, evaluation, and scientific creativity.\\

A complementary line of work focuses on LLM-based agentic reasoning frameworks, which formalize how language models can decompose tasks, call external tools, iteratively refine outputs, and use feedback signals to improve their reasoning trajectories. Recent surveys, including LLM-based Agentic Reasoning Frameworks: A Survey from Methods to Scenarios~\cite{zhao2025llm}, outline the architectural patterns behind contemporary agent systems such as ReAct~\cite{yao2022react}, Tree-of-Thoughts~\cite{yao2023tree}, Chain-of-Thought prompting~\cite{wei2022chain}, and open-ended agents like Voyager~\cite{wang2023voyager}. At the same time, the emergence of scientific LLMs, summarized in A Survey of Scientific Large Language Models~\cite{hu2025survey}, demonstrates a clear trend toward specialized models trained on high-quality scientific corpora, domain-specific reasoning tasks, and structured multimodal inputs. Together, these developments indicate a convergence: scientific LLMs are becoming more agentic, and agentic LLM frameworks are increasingly being adapted to scientific research.\\

Despite these advances, systematic mechanisms for improving agentic scientific reasoning remain underdeveloped, particularly in tasks involving long-horizon decision making such as scientific writing, experimental design, and multi-step argument construction. Existing approaches typically rely on supervised fine-tuning or heuristic feedback rather than grounded optimization principles. Furthermore, progress in this field is hindered by the lack of standardized benchmarks that evaluate the quality, structure, and correctness of scientific text generated by LLMs.\\

To address these gaps, this work makes three contributions. First, we propose a reinforcement learning and value-based training paradigm for scientific paper generation, where the model learns to construct coherent, verifiable, multi-section research manuscripts under explicit value functions. Unlike prior methods that optimize for surface-level textual quality, our framework introduces a structured credit-assignment mechanism tailored to scientific writing. Second, we build a new benchmark for evaluating scientific paper generation, designed around realistic multi-section writing tasks, citation consistency, argumentative structure, and factual validity. This benchmark enables reproducible evaluation and comparative analysis for future work on scientific LLMs and agentic scientific reasoning. Third, our experiments demonstrate that the proposed value-guided generation framework significantly improves global narrative coherence, scientific correctness, and cross-section consistency over strong supervised baselines.

\section{Related Works}

\subsection{AI for Scientific Research and Agentic Reasoning}

Artificial intelligence has become increasingly involved across the scientific research pipeline. Surveys such as AI4Research~\cite{chen2025ai4researchsurveyartificialintelligence} and the evolving role of LLMs in scientific innovation~\cite{zhang2025evolving} document how modern systems now assist with literature review, experiment planning, hypothesis evaluation, and autonomous scientific inquiry. In parallel, research on agentic reasoning explores how LLMs can plan, decompose tasks, use tools, and refine solutions through iterative feedback. The survey by Zhao et al.~\cite{zhao2025llm} organizes this emerging paradigm, and foundational frameworks such as ReAct~\cite{yao2022react}, Tree-of-Thoughts~\cite{yao2023tree}, Chain-of-Thought prompting~\cite{wei2022chain}, and Voyager~\cite{wang2023voyager} illustrate how structured reasoning and action execution can be combined. Multi-agent systems for scientific writing, including SurveyForge~\cite{yan2025surveyforge} and SciSage~\cite{shi2025scisage}, further demonstrate that outline planning, memory-driven generation, and coordinated agent roles improve complex writing tasks. Our work extends this line of research by introducing principled value-based optimization for long-horizon scientific writing.

\subsection{Scientific LLMs and Automated Scientific Writing}

A growing body of work focuses on scientific large language models trained on domain-specific corpora or tailored for scientific reasoning. Hu et al.~\cite{hu2025survey} provide a recent overview, covering data foundations and agentic extensions. Representative models such as Galactica~\cite{taylor2022galactica}, SciGPT~\cite{she2025scigpt}, and SciBERT~\cite{beltagy2019scibert} aim to support scientific retrieval, knowledge grounding, and citation reasoning. Complementing these models, several systems target automatic survey or scientific writing. AutoSurvey~\cite{wang2024autosurvey}, SurveyX~\cite{liang2025surveyx}, and SGSimEval~\cite{guo2025sgsimeval} explore various strategies for survey drafting and multi-dimensional evaluation. More advanced approaches such as SurveyForge~\cite{yan2025surveyforge} and SciSage~\cite{shi2025scisage} employ outline-based heuristics, memory mechanisms, or multi-agent coordination to improve structure and content quality. Recent iterative frameworks for literature analysis and survey automation~\cite{yiran2025analysis,zhang2025deep} highlight the importance of structured intermediate representations and multi-aspect reasoning. Although these methods show strong local improvements, they typically lack a unified optimization principle for controlling global manuscript quality. Our work addresses this limitation by introducing reinforcement learning with explicit value functions to guide multi-section scientific paper generation.

\subsection{Citation Accuracy, Evaluation Benchmarks, and Scientific Text Structure}

Accurate citations are essential in scientific writing and have become a central focus of recent LLM research. ScholarCoPilot~\cite{wang2025scholarcopilot} enhances academic writing through citation-aware triggers and retrieval supervision, achieving substantial improvements in retrieval accuracy and human-judged citation quality. CiteGuard~\cite{choi2025citeguard} complements generation-side methods with retrieval-augmented validation to assess whether model-generated citations faithfully match the claims they support. Beyond citation, a number of benchmarks evaluate scientific text quality from multiple dimensions. SGSimEval~\cite{guo2025sgsimeval} proposes similarity-enhanced metrics for measuring outline quality, content fidelity, and citation behavior. Additional work on scientific text structuring~\cite{bolanos2025modelling,zhu2025context} provides annotated schemas and multi-aspect taxonomies for analyzing scientific discourse. These efforts highlight that scientific writing quality is multi-dimensional, involving structure, factual grounding, and citation faithfulness. Our proposed benchmark draws on these insights, integrating structural, argumentative, and citation-aware criteria while our value-based training framework incorporates them directly into optimization.

\subsection{Reinforcement Learning and Value Modeling for Long-Form Language Generation}

Reinforcement learning has become an influential paradigm for controlling high-level behaviors in large language models, particularly in tasks requiring long-horizon coherence, preference alignment, or structured decision making. Early work on reinforcement learning from human feedback (RLHF) demonstrated that reward-based optimization can significantly improve factuality, helpfulness, and reasoning capabilities in LLMs. Notably, the InstructGPT framework~\cite{ouyang2022training} formalized the modern RLHF pipeline, showing that preference-based optimization can robustly guide model behavior through human demonstrations and comparisons. The underlying optimization is often implemented using Proximal Policy Optimization (PPO)~\cite{schulman2017proximal}, which has become the standard policy-gradient backbone for RLHF systems. More recent extensions, such as reinforcement learning from AI feedback (RLAIF)~\cite{lee2023rlaif} and direct preference optimization (DPO)~\cite{rafailov2023direct}, further expand the feasibility of applying value modeling, preference learning, and policy optimization to long-form generation.

\section{Method}

To more systematically characterize the intermediate reasoning structures underlying scientific paper generation, we reformulate the task from conventional token-level sequence generation into a multi-step evolution process over a structured intermediate state space.

\subsection{Formulation of Scientific Writing}

We formulate scientific paper generation as a structured, long-horizon conditional generation task. Given a set of heterogeneous inputs, the model is required to produce a coherent, well structured scientific article that not only achieves high conditional likelihood, but also satisfies a collection of structural, semantic, and factual constraints implicit in academic writing.

\subsubsection{Input Space}

Let the input be
\begin{equation}
X = (T, C, R),
\end{equation}
where
\begin{itemize}
    \item $T \in \mathcal{T}$ denotes the topic or high level writing intent.
    \item $C = \{c_{1}, c_{2}, \ldots, c_{m}\} \in \mathcal{C}$ represents background materials, documents, or other contextual evidence.
    \item $R = \{r_{1}, r_{2}, \ldots, r_{k}\} \in \mathcal{R}$ is the set of external references available for citation.
\end{itemize}

This formulation is agnostic to how $C$ and $R$ are obtained; in practice, they can come from retrieval modules, pre-defined reading lists, or memory-like components.

\subsubsection{Output Space}

A scientific article is represented as a sequence of textual units:
\begin{equation}
Y = (y_{1}, y_{2}, \ldots, y_{L}),
\end{equation}
where each $y_i \in \mathcal{U}$ may be a token, a phrase, or an element from a domain-specific markup language (e.g., Markdown or \LaTeX). The full output space is defined as
\begin{equation}
\mathcal{Y}
=
\left\{
Y = (y_{1}, \ldots, y_{L})
\;\middle|\;
\begin{array}{l}
L \in \mathbb{N}, \\
y_i \in \mathcal{U}, \\
\mathrm{ValidArticle}(Y)
\end{array}
\right\}.
\end{equation}
where $\mathrm{ValidArticle}(Y)$ denotes that $Y$ conforms to the structural and semantic constraints of a scientific article.\\

During generation, the model parameterized by $\theta$ defines a conditional distribution over $\mathcal{U}$, typically with an autoregressive factorization:
\begin{equation}
\begin{aligned}
\pi_{\theta}(Y \mid X)
&=
\prod_{i=1}^{L}
\pi_{\theta}(y_i \mid y_{<i}, X), \\
&\qquad
\pi_\theta(y_i \mid y_{<i}, X)
\in
\Delta(\mathcal{U}).
\end{aligned}
\end{equation}

\subsubsection{Basic Conditional Generation Objective}

The most general form of the paper-generation task is the maximization of the conditional likelihood:
\begin{equation}
Y^\ast
=
\arg\max_{Y \in \mathcal{Y}}
\pi_{\theta}(Y \mid X),
\end{equation}
where $\pi_{\theta}(Y \mid X)$ denotes the joint likelihood of the entire article under the model. This view recovers standard maximum likelihood training and decoding when no additional constraints are imposed.

\subsubsection{Information Coverage and Structural Constraints}

A scientific article must not only be fluent and coherent but also faithfully utilize information provided in the input materials. Let $I(\cdot)$ denote a semantic information mapping (for example, extracted facts, key claims, or important concepts). The coverage of contextual information is defined as
\begin{equation}
\text{Coverage}(Y, C)
=
\frac{\lvert I(Y) \cap I(C) \rvert}{\lvert I(C) \rvert}.
\end{equation}

Most prior work implicitly or explicitly seeks to maximize this quantity, ensuring that the generated paper incorporates essential content from the input, rather than hallucinating unsupported statements or ignoring core evidence.

The overall objective can be written as a composite optimization problem:
\begin{equation}
Y^\ast
=
\arg\max_{Y \in \mathcal{Y}}
\Big[
\pi_{\theta}(Y \mid X)
+
\sum_{I}
\lambda_{I} \cdot \text{Coverage}(Y, C)
\Big],
\end{equation}
where $\lambda_I$ are tunable coefficients balancing likelihood and coverage-based criteria. This regularized objective captures the intuition that a good scientific article should both be likely under the model and align well with the provided evidence.

\subsubsection{Citation Consistency}

A key requirement of scientific writing is citation correctness. Let $\text{Citation}(Y) \subseteq \mathcal{R}$ denote the set of references cited in the output. A simple citation-consistency objective is given by
\begin{equation}
\text{CitationScore}(Y)
=
\mathbb{1}\big[\text{Citation}(Y) \subseteq \mathcal{R}\big],
\end{equation}
ensuring that all citations correspond to legitimate entries in the provided reference set. More fine-grained variants can additionally measure whether the cited references are relevant to the local context or whether important references from $R$ are missing.

\subsubsection{Constraint-based View}

The above regularized formulation can also be expressed as a constrained optimization problem. Let $\mathcal{S}_{\text{struct}}$, $\mathcal{S}_{\text{fact}}$, and $\mathcal{S}_{\text{cite}}$ denote feasible sets that encode structural validity, factual consistency, and citation constraints, respectively. Then we can write:
\begin{align}
\max_{Y \in \mathcal{Y}} \quad & \pi_{\theta}(Y \mid X) \\
\text{s.t.} \quad
& \text{Coverage}(Y, C) \ge \delta_{\text{cov}}, \\
& Y \in \mathcal{S}_{\text{struct}}, \\
& Y \in \mathcal{S}_{\text{fact}}, \\
& Y \in \mathcal{S}_{\text{cite}},
\end{align}
where $\delta_{\text{cov}}$ controls the desired level of information coverage. In practice, these constraints are often enforced approximately, for example through auxiliary losses, decoding heuristics, or post hoc editing modules.

\subsection{Reinforcement Learning}

Reinforcement learning (RL) provides an alternative perspective to long-form scientific writing by modeling the generation process as a sequential decision-making problem. Instead of optimizing the conditional likelihood directly, RL seeks to maximize a task-specific reward function that captures the desired properties of a scientific article, including coverage, structure, coherence, factuality, and citation correctness.\\

Formally, we consider an episodic Markov decision process (MDP)
\begin{equation}
\mathcal{M} = (\mathcal{S}, \mathcal{A}, P, r, \gamma),
\end{equation}
where $\mathcal{S}$ is the state space, $\mathcal{A}$ is the action space, $P$ is the transition kernel, $r$ is the reward function, and $\gamma \in (0,1]$ is the discount factor. At each step $t$, the model is in state $s_t \in \mathcal{S}$ and emits an action $a_t \in \mathcal{A}$ that corresponds to producing the next textual unit $y_t$, followed by a transition to $s_{t+1}$ and receipt of a scalar reward $r_t$.

In this view, the writing policy $\pi_\theta(a_t \mid s_t)$ defines a distribution over textual actions conditioned on the current partial document. The objective is to maximize the expected return
\begin{equation}
J(\theta) = \mathbb{E}_{\pi_\theta} \Big[ \sum_{t=1}^{L} \gamma^{t-1} r_t \Big],
\end{equation}
where rewards can encode high-level writing desiderata, such as semantic coverage or citation correctness. This formulation decouples training from token-level likelihood and allows integration of non-differentiable constraints through reward shaping.\\

Existing RL approaches for long-form generation often rely on preference modeling, reward models, or heuristic scoring functions. However, directly applying RL to scientific writing is challenging due to extremely long horizons, sparse rewards, and complex structural constraints. These limitations motivate the introduction of hierarchical abstractions and intermediate reasoning states, enabling the model to plan at multiple timescales and reduce the difficulty of credit assignment.

\subsection{Overview and Schema}

Specifically, we represent a paper outline as an editable hierarchical structure, and model the generation process as a sequence of discrete state transitions over outline states $s_{\text{outline}} \in \mathcal{S}$. Each transition is realized through an explicit editing action, denoted as $a_{\text{diff}} \in \mathcal{A}$. Borrowing terminology from code version control systems, we refer to these editing operations as \emph{diffs}. Under this formulation, paper generation is no longer treated as a single-pass sampling of a token sequence, but rather as a reasoning trajectory that progressively refines an intermediate structure: starting from a coarse or empty outline, a sequence of local edits gradually converges to a high-quality outline, which in turn induces the final document generation. The intermediate state space $\mathcal{S}$ can therefore be interpreted as the set of all valid outline structures.\\

To capture human editing preferences, we sample trajectories of outline-editing behaviors from real scientific papers, yielding preference data of the form $(s_{\text{outline}}, a_{\text{diff}}) \sim \mathcal{T}(\mathcal{D}_{\mathrm{Paper}})$. Here, $\mathcal{D}_{\mathrm{Paper}}$ denotes a corpus of raw scientific papers crawled from the arXiv API, and $\mathcal{T}$ is a transcription module that converts document revision histories into state--diff pairs over outline structures. We train a policy $\pi_\theta(a_{\text{diff}, t} \mid s_{\text{outline}, t})$ to align with these human preferences; for brevity, we subsequently write $\pi_\theta(a_t \mid s_t)$.\\

Depending on the intent of an editing diff, some actions (e.g., node reordering, deletion, or structural relocation) can be executed deterministically, while others (e.g., semantic refinement or content expansion) inherently require generative capabilities. For the latter class, the system constructs a query conditioned on the editing intent and delegates execution to a language model. Formally, this process is described as
\begin{equation}
\left\{
\begin{aligned}
& q_t = f(a_t), \\
& s_{t+1} \sim \mathrm{LLM}(\cdot \mid s_t, q_t),
\end{aligned}
\right.
\end{equation}
where $f$ maps an action to a model query.\\

We define an execution-mode indicator function $\kappa: \mathcal{A} \rightarrow \{\texttt{direct}, \texttt{llm}\}$ to distinguish whether a diff can be applied directly or requires language-model-based generation. When $\kappa(a) = \texttt{direct}$, the action corresponds to a structural and deterministic edit; when $\kappa(a) = \texttt{llm}$, the action involves semantic generation or rewriting. This allows us to define a unified state transition operator
\begin{equation}
P(\cdot \mid s, a) \leq
\begin{cases}
\delta\!\left(\mathrm{Exec}(s, a)\right), & \kappa(a) = \texttt{direct}, \\
\mathrm{LLM}(\cdot \mid s, f(a)), & \kappa(a) = \texttt{llm},
\end{cases}
\end{equation}
where $\mathrm{Exec}(s, a)$ is a deterministic structural editing function and $\delta(\cdot)$ denotes a degenerate (deterministic) distribution.\\

Under this formulation, the outline generation process follows $s_{t+1} \sim P(\cdot \mid s_t, a_t)$ and can be formalized as a Markov Decision Process (MDP) $\mathcal{M} = \langle \mathcal{S}, \mathcal{A}, P, r, \gamma \rangle$. A complete generation trajectory is defined as $\tau = (s_0, a_0, s_1, a_1, \dots, s_T)$, where the initial state $s_0$ corresponds to an empty outline. Subsequent states are generated through the alternating interaction between the policy $\pi_\theta$ and the transition operator $P$:
\begin{equation}
a_t \sim \pi_\theta(\cdot \mid s_t), \quad
s_{t+1} \sim P(\cdot \mid s_t, a_t).
\end{equation}

Human preference data can thus be interpreted as trajectories induced by an unknown optimal policy $\pi^\ast$, i.e., $\tau \sim p_{\pi^\ast}(\tau)$. We learn a parameterized policy $\pi_\theta$ via maximum likelihood estimation or preference alignment objectives to approximate this distribution, enabling the model to reproduce human editing decisions within the structured outline-editing space. Importantly, the learned policy not only produces high-quality outline structures, but also explicitly exposes intermediate reasoning steps and structural evolution trajectories, yielding an interpretable and controllable representation.

\subsection{Methodology for Benchmarking}

Our benchmarking data are constructed by systematically parsing research papers from arXiv. In this work, we sample $1,500$ articles spanning six high-level research domains and eleven fine-grained subcategories. While the experiments in this paper are conducted on a fixed subset for analysis and reproducibility, it is important to note that our proposed pipeline is fully scalable and can be applied to an arbitrarily large collection of arXiv articles to continuously generate training data.\\

Inspired by how abstract syntax trees (ASTs) are extracted and manipulated in code-centric tasks, we parse the structural organization of scientific papers through their arXiv HTML representations. This allows us to recover a hierarchical document schema consisting of sections, subsections, and paragraphs. Based on this structured representation, we employ an augmentation model to propose potential modification operations over the document content.\\

Concretely, we sample an individual paragraph and apply stochastic perturbations, such as deleting or modifying a subset of its content. The resulting corrupted paragraph is treated as a problem instance, while the target solution corresponds to reasoning about a coherent and appropriate modification plan that restores or improves the paragraph. In this way, a planning problem over document evolution is reformulated as a pure reasoning task. Each instance yields a pair of states before and after modification, together with the corresponding modification prompt. Notably, this process mirrors the inverse direction of our document generation pipeline, enabling the model to learn how local edits accumulate into global structural improvements.

\begin{figure}[t]
    \centering
    \begin{minipage}[t]{0.48\textwidth}
        \centering
        \includegraphics[width=\linewidth]{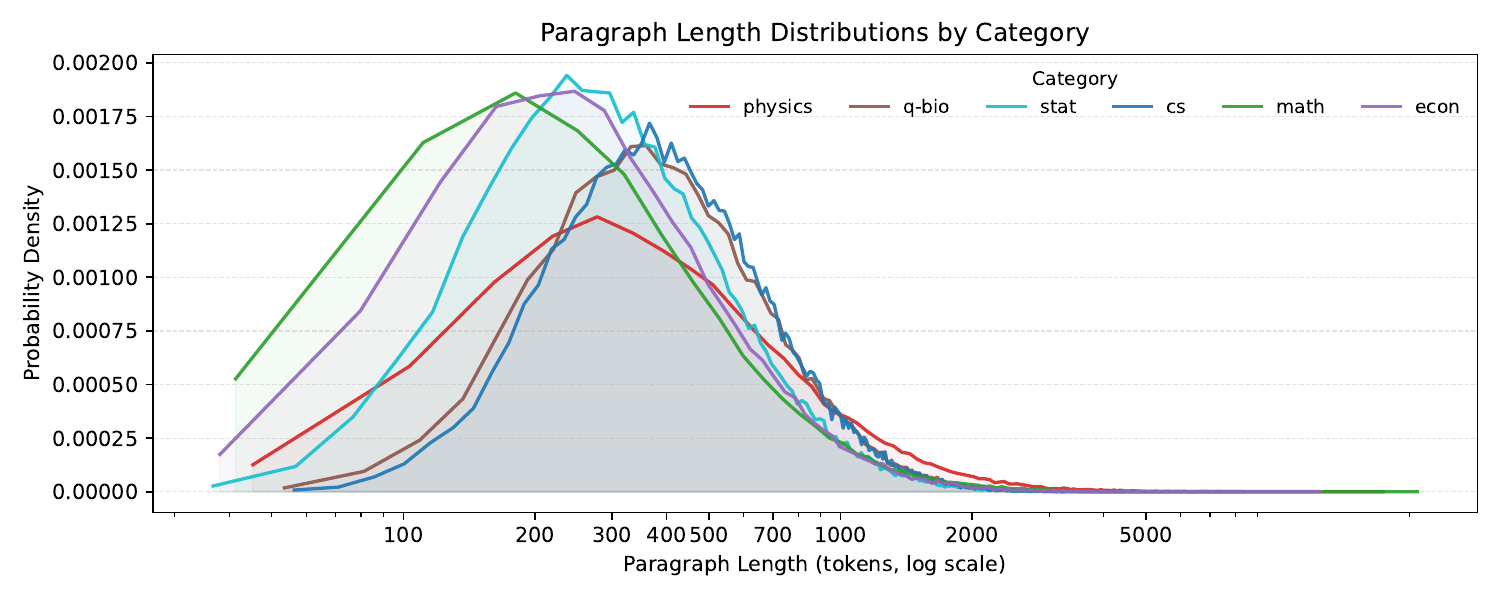}
        \vspace{0.8em}
        \includegraphics[width=\linewidth]{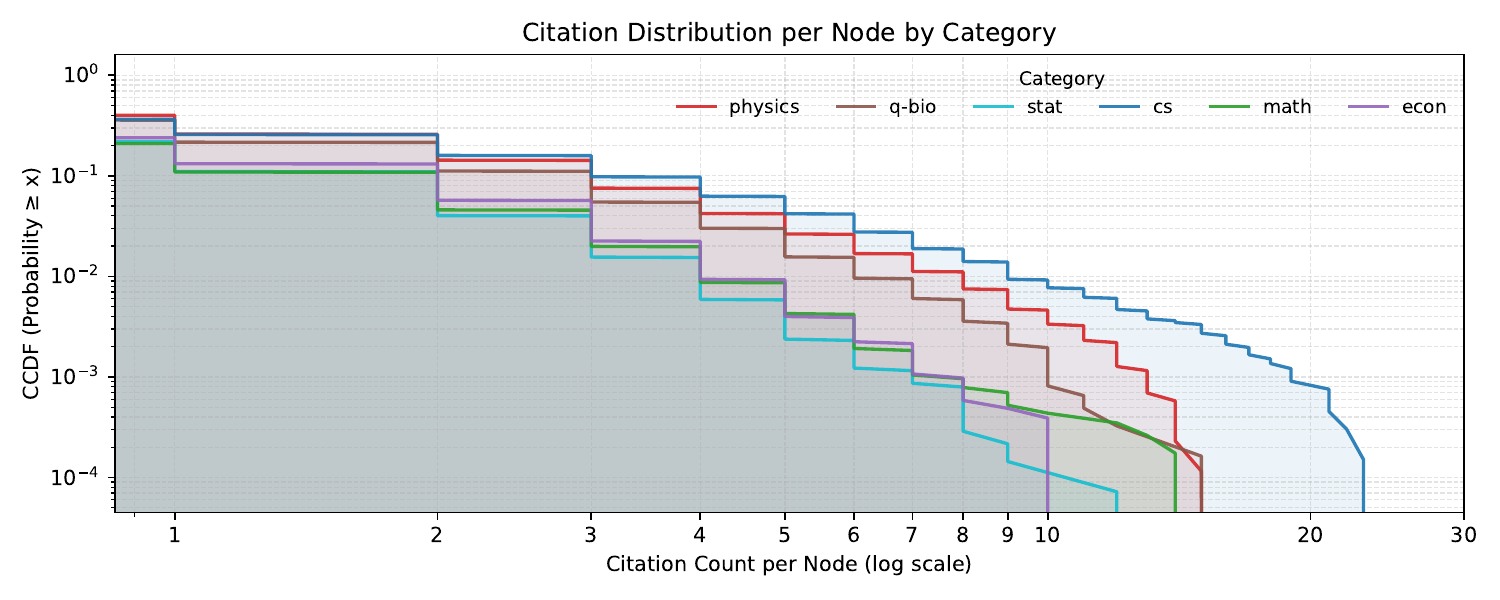}
    \end{minipage}
    \hfill
    \caption{Paragraph length and citation density distributions by category.}
    \label{fig:distribution_analysis}
\end{figure}

Figure~\ref{fig:distribution_analysis} summarizes key statistical properties of the constructed dataset. Specifically, it reports the distribution of individual paragraph lengths and the citation density per paragraph across major research categories. A notable observation is that the citation distribution is highly skewed rather than uniform. This characteristic helps explain the limitations of prior approaches, many of which rely on retrieving a fixed number of references at each step via retrieval-augmented generation (RAG). In practice, however, the true citation demand varies substantially across different stages of document development. For clarity, we present category-level averages instead of fine-grained subcategory statistics, as the latter exhibit similar trends and do not materially affect the conclusions.\\

\begin{figure}[t]
    \centering
    \begin{minipage}[t]{0.48\textwidth}
        \hfill
        \includegraphics[width=\linewidth]{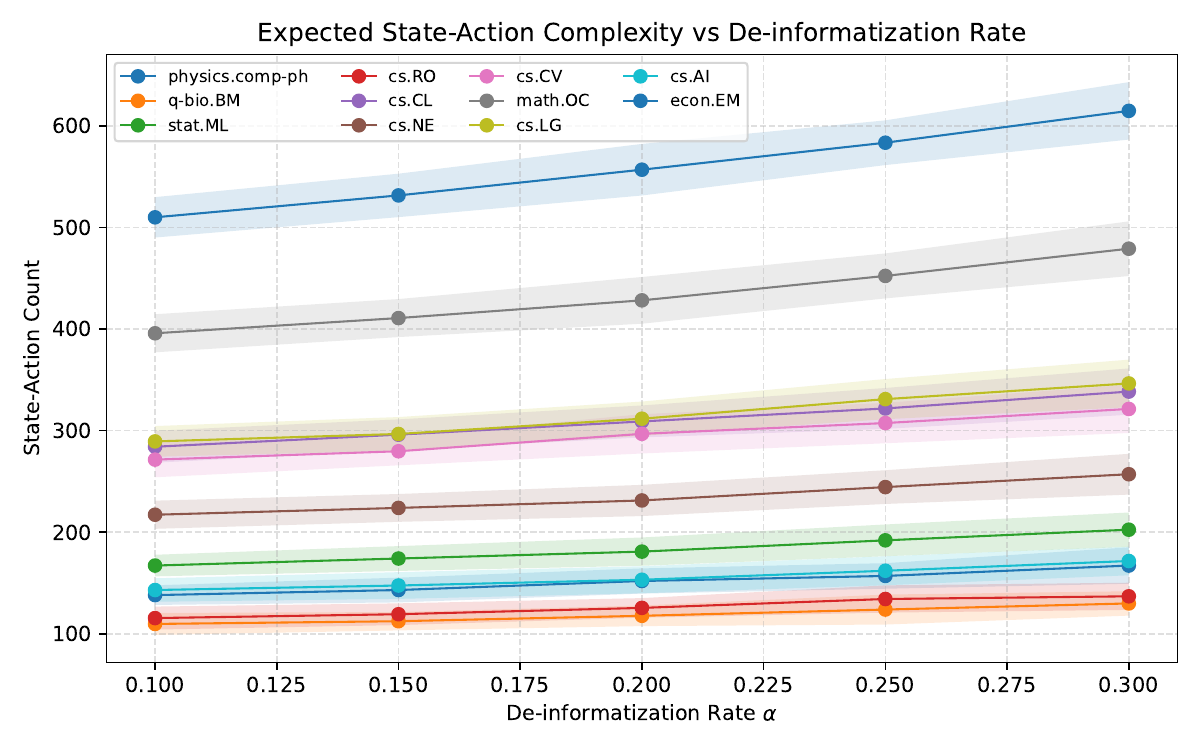}
    \end{minipage}
    \caption{State-action complexity under varying de-informatiation rates.}
    \label{fig:state_action_dynamics}
\end{figure}

Figure~\ref{fig:state_action_dynamics} analyzes the expected number of steps required to fully decompose a complete article into an empty outline, which serves as a proxy for estimating the planning horizon in document evolution. The x-axis represents the proportion of update operations, which preserve the document structure but are nevertheless crucial for producing high-quality articles. Empirically, we find that for the majority of research domains, a complete article can be generated within 200 to 300 steps. Based on this observation, we adopt these two step budgets as the primary experimental settings in subsequent evaluations.

\section{Experiments}

We evaluate our approach on a survey generation task, which is closely aligned with prior systems such as SurveyForge, SciSage, and AutoSurvey. 
Survey generation serves as a representative long-horizon scientific writing task that requires coherent structure planning, iterative content refinement, and effective reference utilization. 
Unless otherwise specified, all experiments are conducted under two inference budgets, 200 steps and 300 steps, which are selected based on the empirical analysis presented in Section~3. \\

We fine-tune three compact models, namely Gemma2-2B, Qwen3-1.8B, and Phi-3.8B, using the training data generated by our proposed pipeline. 
In addition, we evaluate several external large language models in a zero- or few-shot manner, including GPT-4o-mini, Claude-3.5-Haiku, and Llama-3.1-Instruct-70B. 
This setup allows us to examine both the effect of model scale and the benefit of task-specific fine-tuning under our data construction paradigm. \\

\begin{table*}[ht]
\centering
\small
\renewcommand{\arraystretch}{1.2}
\begin{tabular}{lcccccc}
\hline
\multirow{2}{*}{Model} & \multicolumn{3}{c}{200 steps} & \multicolumn{3}{c}{300 steps} \\
\cline{2-7}
 & Precision & Recall & F1 & Precision & Recall & F1 \\
\hline
\textbf{Ours (GPT-4o-mini)} & 0.612$\pm$0.089 & 0.264$\pm$0.105 & 0.352$\pm$0.083 & 0.668$\pm$0.103 & 0.312$\pm$0.151 & 0.397$\pm$0.164 \\
SurveyForge & 0.346$\pm$0.028 & 0.261$\pm$0.113 & 0.285$\pm$0.083 & \textemdash & \textemdash & \textemdash \\
AutoSurvey  & 0.295$\pm$0.028 & 0.181$\pm$0.080 & 0.213$\pm$0.062 & \textemdash & \textemdash & \textemdash \\
\hline
\textbf{Ours (Claude-3.5-Haiku)} & 0.533$\pm$0.094 & 0.173$\pm$0.071 & 0.249$\pm$0.082 & 0.524$\pm$0.069 & 0.186$\pm$0.129 & 0.246$\pm$0.119 \\
SurveyForge & 0.267$\pm$0.027 & 0.265$\pm$0.162 & 0.243$\pm$0.082 & \textemdash & \textemdash & \textemdash \\
AutoSurvey  & 0.228$\pm$0.027 & 0.172$\pm$0.073 & 0.187$\pm$0.050 & \textemdash & \textemdash & \textemdash \\
\hline
\textbf{Ours (Phi-3.8B, Finetuned)} & 0.678$\pm$0.081 & 0.317$\pm$0.116 & 0.422$\pm$0.117 & 0.717$\pm$0.126 & 0.311$\pm$0.225 & 0.381$\pm$0.206 \\
SurveyForge & 0.359$\pm$0.028 & 0.345$\pm$0.274 & 0.313$\pm$0.129 & \textemdash & \textemdash & \textemdash \\
AutoSurvey  & 0.296$\pm$0.029 & 0.306$\pm$0.198 & 0.274$\pm$0.101 & \textemdash & \textemdash & \textemdash \\
\hline
\textbf{Ours (Llama-3.1-Instruct-70B)} & 0.446$\pm$0.064 & 0.161$\pm$0.078 & 0.223$\pm$0.076 & 0.512$\pm$0.130 & 0.196$\pm$0.224 & 0.210$\pm$0.153 \\
\hline
\textbf{Ours (Gemma2-2B, Finetuned)} & 0.399$\pm$0.108 & 0.166$\pm$0.134 & 0.201$\pm$0.112 & 0.406$\pm$0.109 & 0.251$\pm$0.282 & 0.247$\pm$0.175 \\
\hline
\textbf{Ours (Qwen3-1.8B, Finetuned)} & 0.244$\pm$0.084 & 0.095$\pm$0.071 & 0.120$\pm$0.070 & 0.298$\pm$0.126 & 0.074$\pm$0.052 & 0.111$\pm$0.070 \\
\hline
\end{tabular}
\centering
\caption{erformance under different prompt ratios and inference steps. All metrics are reported as percentages.}
\label{tab:model_results}
\end{table*}

The quantitative results are summarized in Table~\ref{tab:model_results}. 
Overall, we observe that prior knowledge plays a critical role in survey generation performance. 
Models below the 2B parameter scale exhibit limited gains after fine-tuning, suggesting that insufficient intrinsic world knowledge constrains their ability to benefit from structured document evolution signals. 
In contrast, once the model capacity reaches a moderate scale, fine-tuning on our data consistently improves performance and, in several cases, surpasses strong general-purpose large models that are not adapted to this task. \\

SurveyForge and AutoSurvey generate survey outlines in a single pass and therefore do not operate under a step-based formulation. 
For completeness, we include their reported performance as reference baselines, although their generation paradigm is fundamentally different from our iterative evolution framework. 
This comparison highlights the advantage of modeling scientific writing as a sequential decision process rather than a one-shot generation problem. \\

While our method is not limited to survey generation and can be applied to general scientific paper writing, we do not include additional task-specific benchmarks due to the lack of standardized evaluation protocols. 
Nevertheless, our state-action formulation allows arbitrary modification prompts to be injected during generation, providing a flexible mechanism for guiding document evolution beyond survey-style outputs. 
We leave the systematic evaluation of these broader applications to future work. \\

\begin{figure*}[htbp]
    \centering

    \begin{subfigure}[b]{0.32\textwidth}
        \includegraphics[width=\linewidth]{{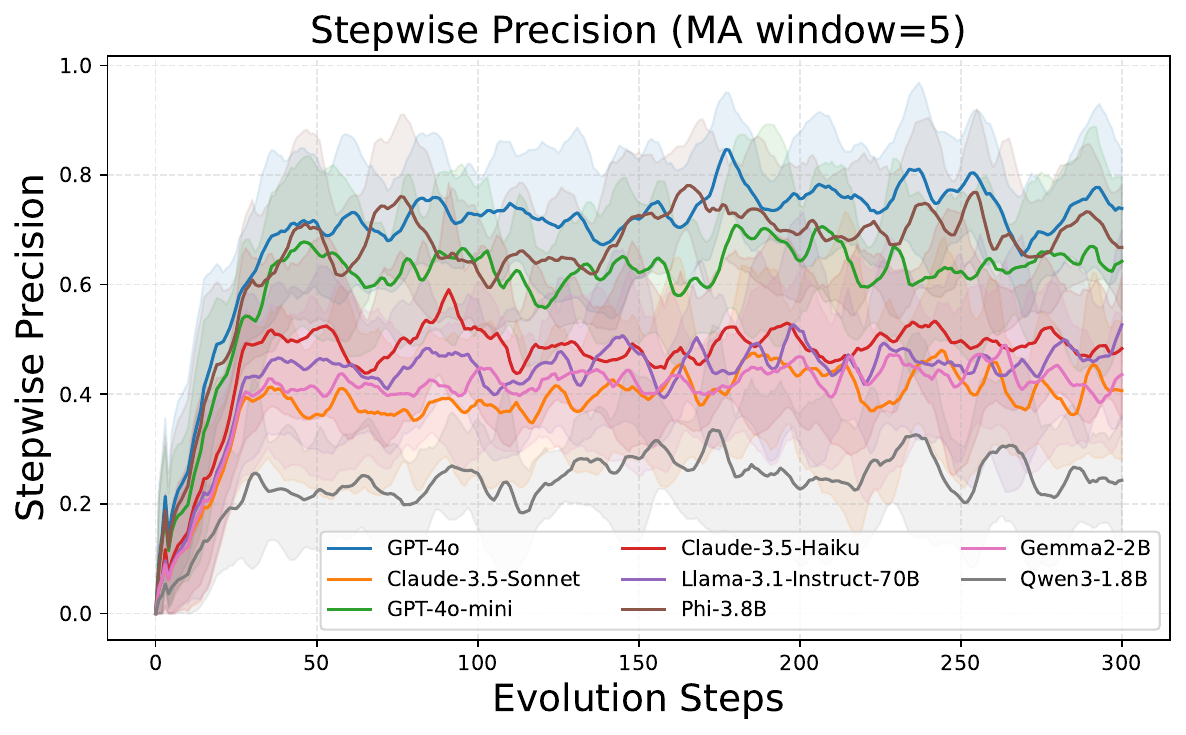}}
    \end{subfigure}
    \hfill
    \begin{subfigure}[b]{0.32\textwidth}
        \includegraphics[width=\linewidth]{{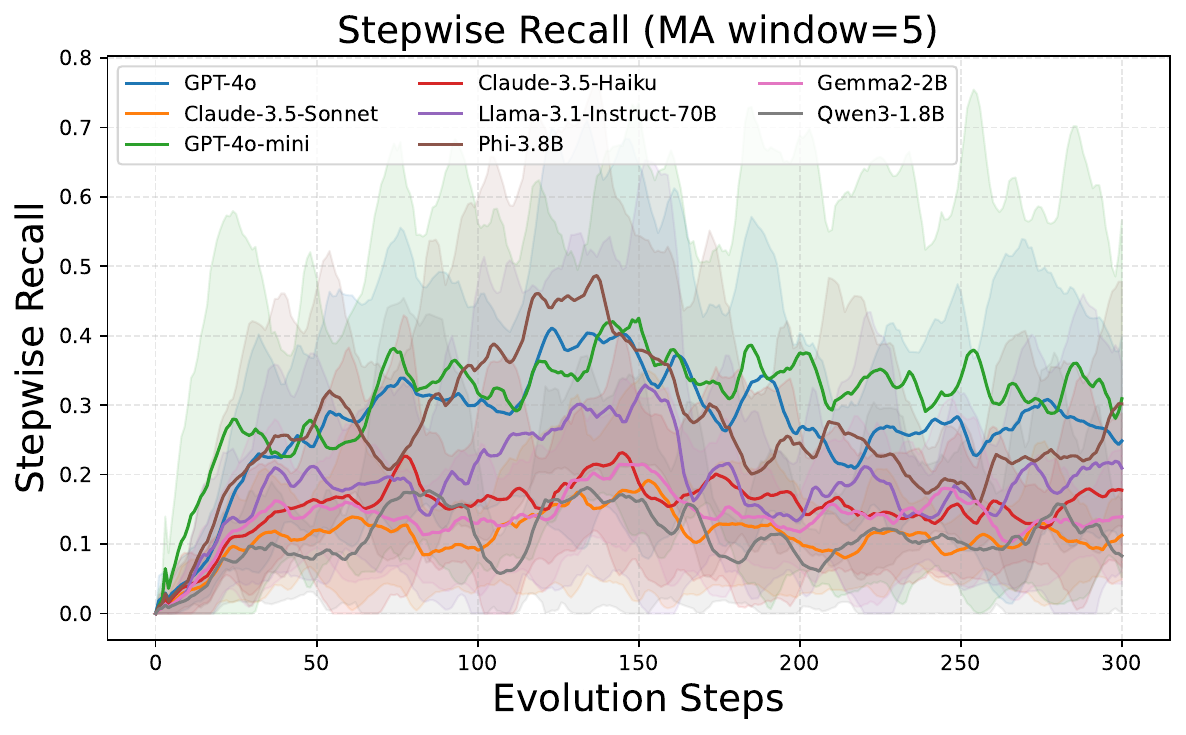}}
    \end{subfigure}
    \hfill
    \begin{subfigure}[b]{0.32\textwidth}
        \includegraphics[width=\linewidth]{{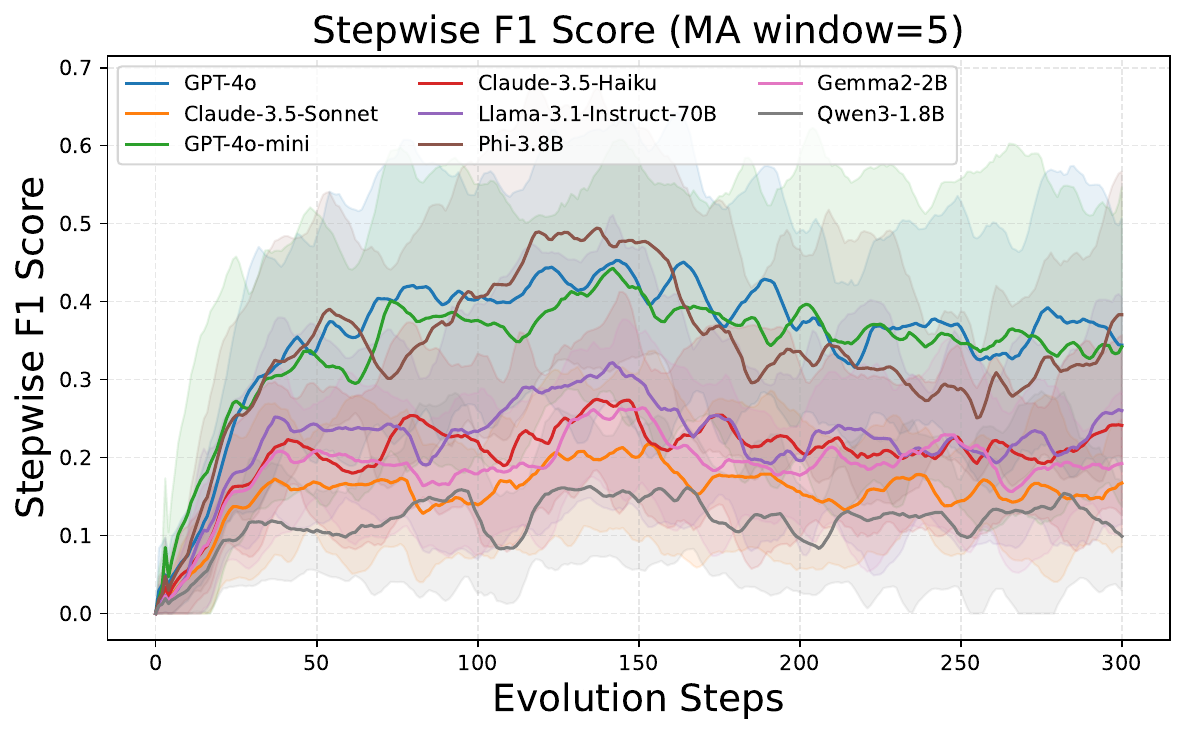}}
    \end{subfigure}

    \vspace{0.5em}

    \begin{subfigure}[b]{0.32\textwidth}
        \includegraphics[width=\linewidth]{{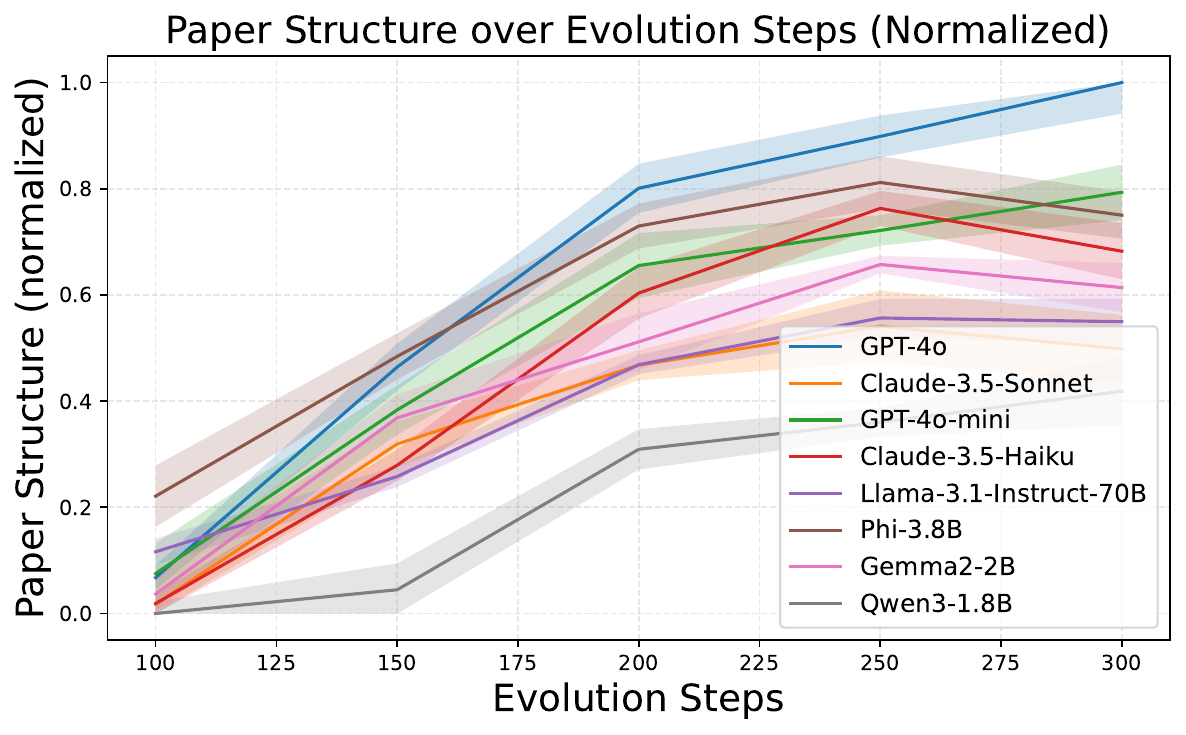}}
    \end{subfigure}
    \hfill
    \begin{subfigure}[b]{0.32\textwidth}
        \includegraphics[width=\linewidth]{{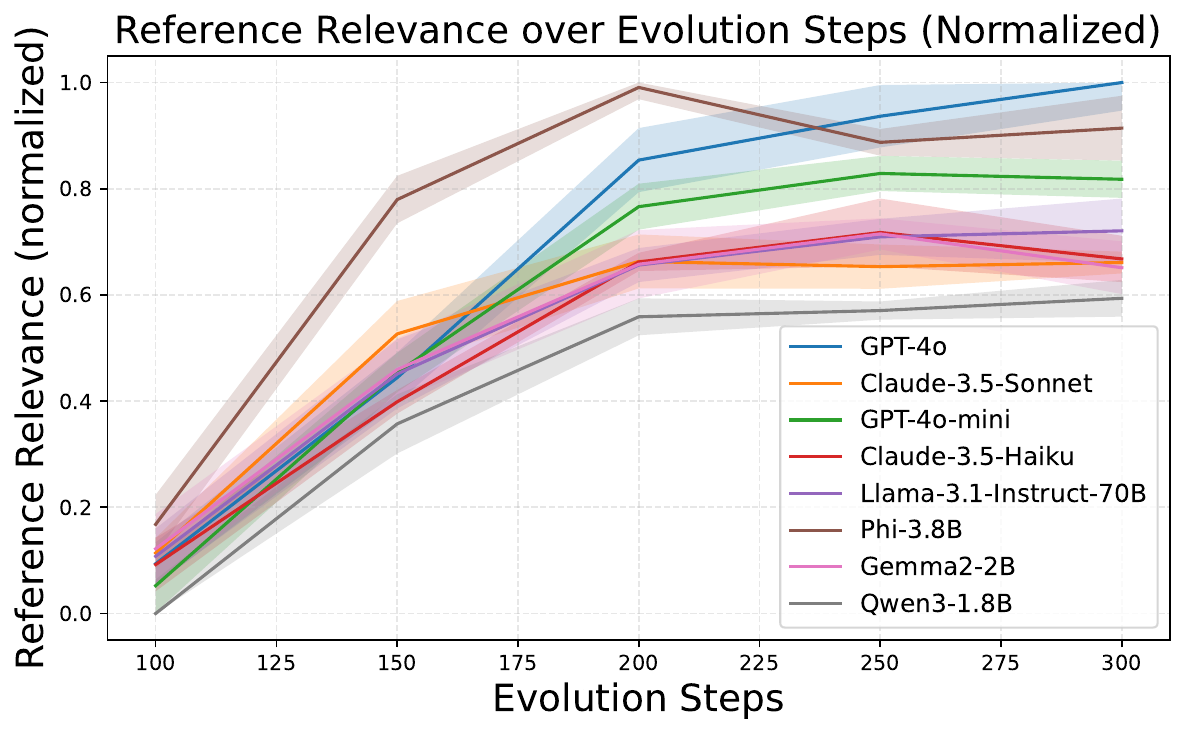}}
    \end{subfigure}
    \hfill
    \begin{subfigure}[b]{0.32\textwidth}
        \includegraphics[width=\linewidth]{{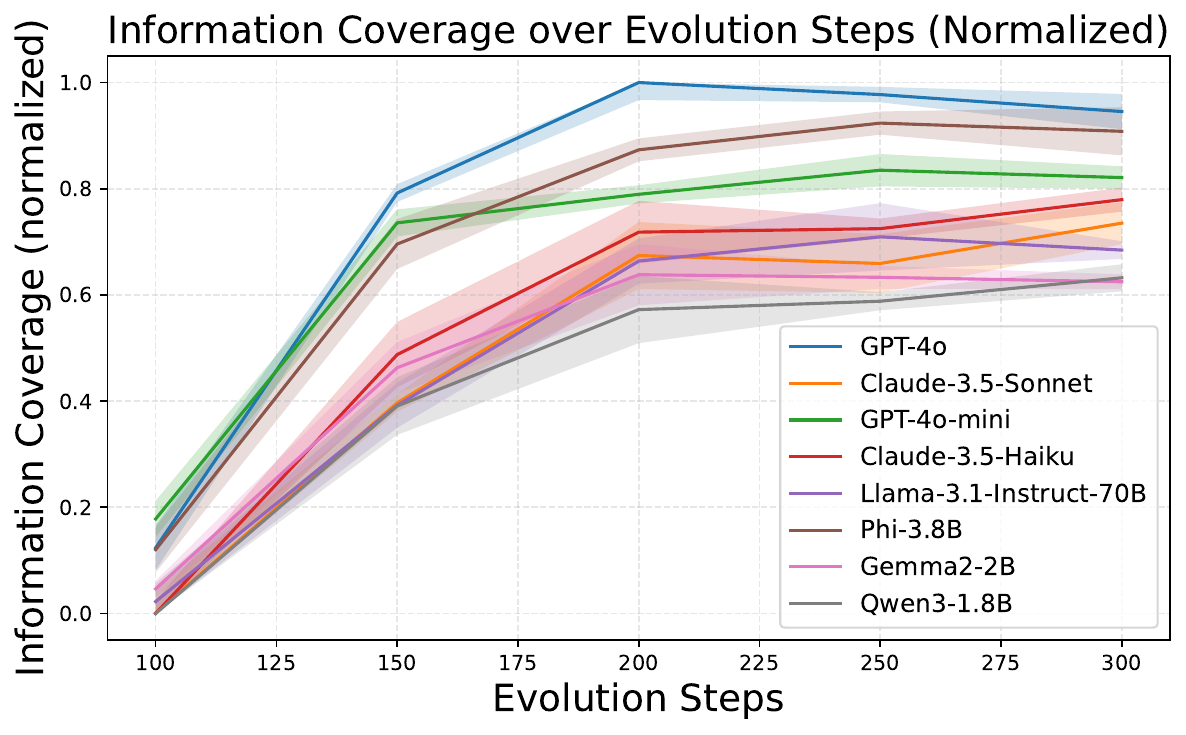}}
    \end{subfigure}

    \caption{Overall caption for the 3x2 figure.}
    \label{fig:3x2}
\end{figure*}

To further analyze the dynamics of the evolution process, we conduct an additional experiment illustrated in Fig.~\ref{fig:3x2}. 
The first three plots report the averaged results over multiple survey generation runs, showing that reference retrieval quickly saturates within the first 50 to 150 steps. 
Beyond this point, the number and relevance of retrieved citations exhibit marginal improvement, indicating that the document structure has largely stabilized. \\

The remaining three plots in Fig.~\ref{fig:3x2} present evaluations of structural completeness, citation relevance, and information density, as assessed by GPT-4o. 
Despite the plateau in citation-related metrics, these higher-level quality measures continue to improve in later stages of generation, suggesting that subsequent steps primarily focus on refinement rather than structural expansion. 
Due to the high cost of automated evaluation, these metrics are computed at a coarse granularity of once every 50 steps. \\

\section{Limitations}

Despite the promising results, our work has several limitations that are worth discussing. First, the evaluation of long-form scientific writing remains inherently challenging, and part of our analysis relies on large language models as automatic judges to assess structural completeness, citation relevance, and information density. Although this practice is increasingly common, such evaluations may introduce biases aligned with the judging models' own priors and preferences, and may not fully reflect human expert judgment. \\

Second, our experimental validation primarily focuses on the survey generation task, which, while representative of structured scientific writing, does not cover the full spectrum of academic writing scenarios such as original research articles, methodological papers, or interdisciplinary works. Extending the evaluation to more diverse scientific writing tasks would provide a more comprehensive understanding of the generality of our approach. \\

Third, the state and action spaces used in our framework are defined based on manually designed schemas derived from document structures and editing operations. While this design offers interpretability and controllability, it may limit flexibility when adapting to domains with substantially different writing conventions or document formats. Automatically learning or refining state-action abstractions remains an open challenge. \\

Finally, although our method supports long-horizon planning, error accumulation across iterative editing steps may still occur, particularly when early-stage structural decisions are suboptimal. Investigating mechanisms for global revision, rollback, or hierarchical planning could further improve robustness in long-generation regimes. \\

\section{Conclusion}

In this work, we propose a state-action-based framework for modeling long-horizon scientific writing as an iterative planning and reasoning process. By inverting document edits into structured supervision signals, we transform document evolution into a tractable reasoning problem that can be used to generate scalable training data. Our analysis of arXiv-derived statistics further reveals structural properties of scientific writing, such as highly skewed citation distributions, that are often overlooked by existing methods. \\

Experimental results on survey generation demonstrate that our approach consistently outperforms strong baselines and enables smaller, fine-tuned models to surpass larger general-purpose models under comparable settings. Beyond survey generation, our framework naturally supports flexible editing objectives and non-survey scientific writing, highlighting its potential as a general paradigm for controllable long-form text generation. We hope this work will inspire future research on self-evolving systems and structured reasoning for complex generative tasks. \\

\bibliography{custom}

\appendix



\end{document}